%% file: ms.tex
\documentclass[10pt,conference,letter,twocolumn]{IEEEtran} 
\usepackage[utf8]{inputenc}
\usepackage[english]{babel}
\usepackage{authblk}

\usepackage{graphicx}
\usepackage{subfig}
\usepackage{comment}
\usepackage{geometry}
\usepackage{balance}
\usepackage{biblatex}
\usepackage{color}

\input{MySetting}

\title{CholecSeg8k: A Semantic Segmentation Dataset for Laparoscopic Cholecystectomy Based on Cholec80}

\author{W.-Y. Hong, C.-L. Kao, Y.-H. Kuo, J.-R. Wang, W.-L. Chang and C.-S. Shih}
\affil{Graduate Institute of Networking and Multimedia\protect\\ Department of Computer Science and Information Engineering\protect\\ National Taiwan University, Taipei, Taiwan, 106}
\affil{\texttt{cshih@csie.ntu.edu.tw}}
\date{October 2020}

\twocolumn
\begin{document}
\maketitle

\input{00Abstract}
\input{01Intro}

\input{02Description}
\input{03Example}
\input{04Access}
\printbibliography
\balance
\end{document}

%% file: MySetting.tex
\def\slug{\hbox to 6pt{\hfill}\hfill\llap{\vrule height 6pt width 6pt depth 0pt}}
\def\proof{\futurelet\next{\bf Proof:}}

\def\appear#1{\def\@appear{#1}}
\appear{Conference of Fun}

\newcount\timehh\newcount\timemm
\timehh=\time \divide\timehh by 60
\timemm=\time \count255=\timehh \multiply\count255 by -60 \advance\timemm by \count255
\def\paperdraftheadertime{Draft of \today\ at \ifnum\timehh<10 0\fi\number\timehh\,:\,\ifnum\timemm<10 0\fi\number\timemm}%

\def\paperdraftheaderday{\footnotesize Draft of \today\ for \@appear.\ }%

\def\submission#1{\def\@submission{#1}}
\submission{Conference of Fun}

\def\submitheaderday{\footnotesize Submitted to \@appear.\  }%

\def\appearinheaderday{\footnotesize Appeared in \@appear.\  }

\usepackage{color,soul}
\usepackage{myColor} 
\input{mycolor.cfg}

\usepackage[colorinlistoftodos]{todonotes}
\newcounter{mycomment}

%% file: 00Abstract.tex
\begin{abstract}
Computer-assisted surgery has been developed to enhance surgery
correctness and safety. However, the researchers and engineers suffer
from limited annotated data to develop and train better
algorithms. Consequently, the development of fundamental algorithms
such as Simultaneous Localization and Mapping (SLAM) are limited. This
article elaborates the efforts of preparing the dataset for semantic
segmentation, which is the foundation of many computer-assisted
surgery mechanisms. Based on the Cholec80 dataset~\cite{cholec80}, we
extracted 8,080 laparoscopic cholecystectomy image frames from 17
video clips in Cholec80 and annotated the images. The dataset is named
{\bf CholecSeg8K} and its total size is 3GB. Each of 
these images are annotated at pixel-level for thirteen classes, which
are commonly founded in laparoscopic cholecystectomy surgery.
CholecSeg8k is released under the license CC BY-NC-SA 4.0.
\end{abstract}

%% file: 01Intro.tex
\section{Introduction}

Endoscopy is a common procedure for detection, diagnosis, and
treatment on organs, which are difficult to examine without surgery,
such as esophagus, stomach, and colon. In a clinical setting, an
endoscopist navigates an endoscope by controlling the handle while
watching the output video simultaneously recorded on an external
monitor. However, the outcome of endoscopy procedures are
highly dependent on the operator’s training and proficiency
level. Unsteady hand-control motion and organ motion could
significantly affect the outcomes of image analysis and operations.

Many computer-assisted systems have been developed to facilitate
surgeon conducting endoscopic surgeries by providing guidance and
additional context-specific information to the surgeon during the
operation~\cite{Munzer}. Several of such systems use magnetic or
radio-based external sensors to estimate and detect the endoscope
location within the patient’s body. However, these are prone to
errors and it is not practical to achieve sub-centimeter localization
accuracy. For example, the system proposed by Hu {\it et
  al.}~\cite{MagnetLocalization05} requires 16 sensors to achieve 5.6
mm localization accuracy.

Simultaneous Localization and Mapping (SLAM) is an image-based
approach for localizing the camera at pixel level accuracy and
requires no infrastructure. This method yields a real-time 3D map by
comparing sensed data and reference data, and provides an estimate of
the camera location in the 3D space. Fortunately, the reference data
can be collected in advance. For instance, 256 beam LiDAR and GPS-RTK
are used to collect high-definition point cloud maps in
advance. During the runtime, the others can use 16 beams LiDAR to
collect point cloud data and localize itself at 5 centimeter
accuracy. To apply SLAM to endoscope navigation, an accurate
assessment of semantic segmentation on images are inevitable.

One of key requirements for accurate SLAM is to use the correct
reference data to compare with. In an outdoor environment, one can use
GPS to obtain location at large and query for the corresponding
reference data, indexed by GPS location. However, this method is not
practical for endoscopy. One possible approach is to identify the
organs shown in the images and, then, uses the results to query for
the reference data. The first part can be conducted by either semantic
segmentation, which labels object class for each pixel, or object
detection, which labels object class for a bound box. Both semantic
segmentation and object detection require well-labelled image data to
train the prediction networks, which are {\it not} widely available
for endoscope images.

\begin{figure*}[ht]
\centering
    \subfloat[Example Endoscope Image]{
      \includegraphics[width=0.95\columnwidth]{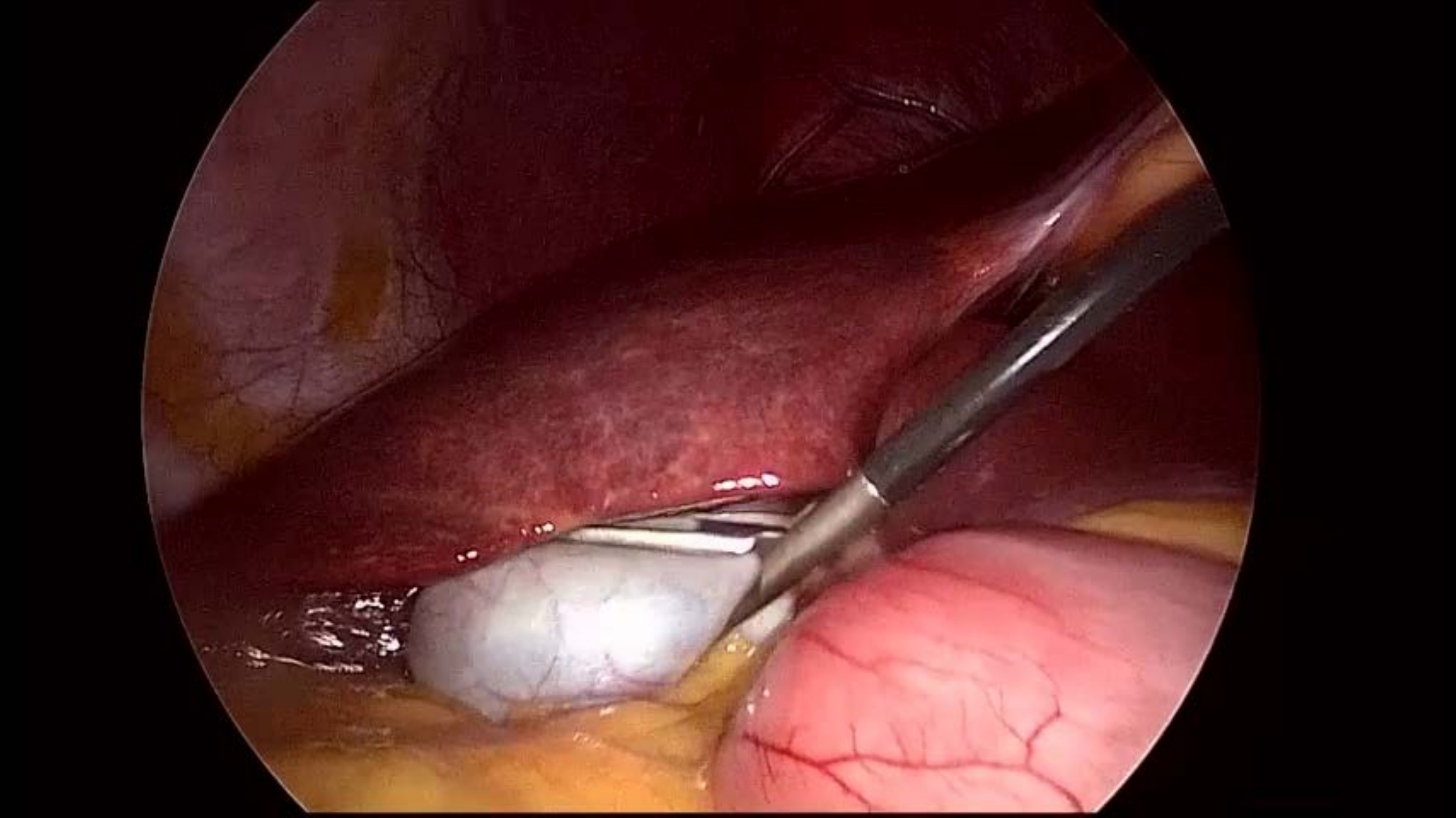}}
    ~
    \subfloat[Example Label]{
        \includegraphics[width=0.95\columnwidth]{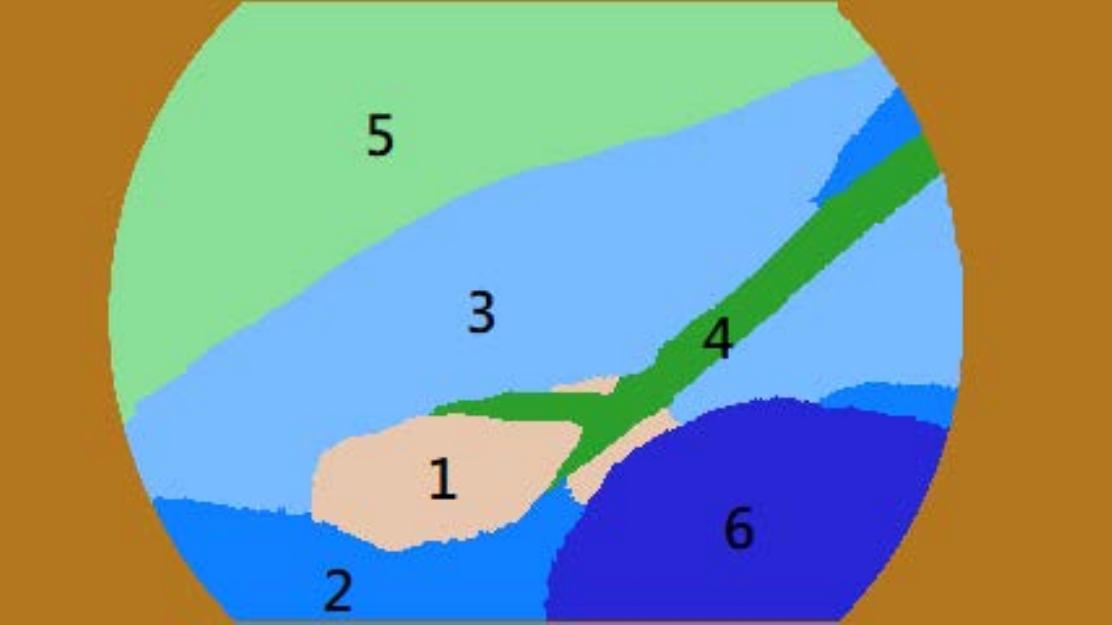}}
    ~
    \caption{Example of Semantic Segmentation Label of Endoscope Image}
    \label{fig:ExampleEndoscopeImage}
\end{figure*}

We construct an open semantic segmentation endoscopic dataset, which
is available to medical and computer vision communities. The
researchers can eliminate the efforts on training the prediction
models, such as FCN, PSPNet, U-Net, and other machine learning
approaches, e.g., Deeplab series. This dataset consists of in total
8,080 frames extracted from 17 video clips in Cholec80
dataset~\cite{cholec80, endonet}. The annotation has thirteen
classes.
Figure~\ref{fig:ExampleEndoscopeImage} shows one example of the image and annotations. 
Figure~\ref{fig:ExampleEndoscopeImage}(a) is the raw image data from Cholec80 dataset and Figure~\ref{fig:ExampleEndoscopeImage}(b) is the color mask to represent the class of the pixels in the image. The details of the annotation will be presented later.

%% file: 02Description.tex
\section{Description of CholecSeg8K Dataset}

CholecSeg8K dataset uses the endoscopic images from
Cholec80~\cite{cholec80}, which is published by CAMMA (Computational
Analysis and Modeling of Medical Activities) research group, as the
base. The research group cooperated with the University Hospital of
Strasbourg, IHU Strasbourg, and IRCAD to construct the
dataset. Cholec80 contains 80 videos of cholecystectomy surgeries
performed by 13 surgeons. Each video in Cholec80 dataset recorded the
operations at 25 fps and has the annotation for instruments and
operation phases. Our work selected a subset of the highly related
videos from Cholec80 dataset and annotated semantic segmentation masks
on extracted frames in the selected videos.

\begin{figure}[htbp]
	\centering
        \includegraphics[width=0.4\columnwidth]{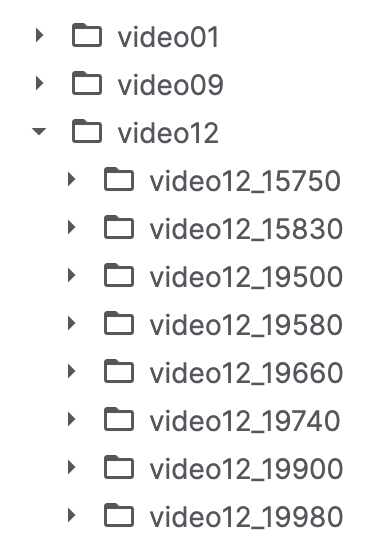}
    \caption{Example of Directory Tree}
	\label{fig:Directory}
\end{figure}

Data in CholecSeg8K Dataset are grouped into a two-level directory for
better organization and accessibility. Figure~\ref{fig:Directory}
shows part of the directory for elaboration. Each directory on the
first level collect the data of the video clips extract from Cholec80
and is named by the filename of the video clips. As examples, {\tt
  video01} and {\tt video09} in Figure~\ref{fig:Directory} are two
different video clips, i.e., {\tt 01} and {\tt 09}, in Cholec80
dataset. Each directory on the secondary level tree stores the data
for 80 images from the video clip and is named by the video filename
and the frame index of the first image in the selected video clip. As
examples, directory {\tt video12\_15750} stores the data from the
$15750$-th frame to the $15829$-th frame from video file {\tt
  video12}. Each secondary level directory stores the raw image data,
annotation, and color masks for 80 frames. There are a total of 101
directories and the total number of frames is 8,080. The resolution of
each image is $854\mbox{\,pixels}\times480\mbox{\,pixels}$. The total
size of the data set is 3GB.

The reason for not annotating all the frames is to skip the operations
not related to semantic segmentation. For example, preparation phase
and ending phase are not annotated.

The number of annotation classes in this dataset is 13,
including black background, abdominal wall, liver, gastrointestinal tract, fat, grasper, connective tissue, blood,
cystic duct, L-hook electrocautery (Instrument), gallbladder, 
hepatic vein, and liver ligament. Table~\ref{tab:class_table} shows the corresponding class names of the class ID in the dataset. 
\begin{center}
\begin{table}[h]
\centering
\begin{tabular}{|c|c|}
    \hline
     Class ID &  Class Name\\
     \hline
     Class 0 & Black Background\\
     \hline
     Class 1 & Abdominal Wall\\
     \hline
     Class 2 & Liver\\
     \hline
     Class 3 & Gastrointestinal Tract\\
     \hline
     Class 4 & Fat\\
     \hline
     Class 5 & Grasper\\
     \hline
     Class 6 & Connective Tissue\\
     \hline
     Class 7 & Blood\\
     \hline
     Class 8 & Cystic Duct\\
     \hline
     Class 9 & L-hook Electrocautery\\
     \hline
     Class 10 & Gallbladder\\
     \hline
     Class 11 & Hepatic Vein\\
     \hline
     Class 12 & Liver Ligament\\
     \hline
\end{tabular}
\caption{Class numbers and their corresponding class names}
\label{tab:class_table}
\end{table}
\end{center}

While annotating the pixels, the annotation classes are defined to aim
on cholecystectomy surgeries, which is the targeted operations in
Cholec80 dataset. Specifically, the annotation classes aim to
recognize liver and gallbladder. Other annotation classes, which are
not closely related to the surgeries, are defined for broader
coverage. In particular, two classes have broader coverage.  They are
{\em gastrointestinal tract} and {\em liver ligament}. The
gastrointestinal tract class includes stomach, small intestine, and
nearby tissues. The liver ligament class includes coronary ligament,
triangular ligament, falciform ligament, ligamentum teres (hepatis),
ligamentum venosum, and lesser omentum.

\begin{figure}[hbt!]
  \centering
  \subfloat[ht][Group with larger proportion]{
    \includegraphics[width=0.9\columnwidth]{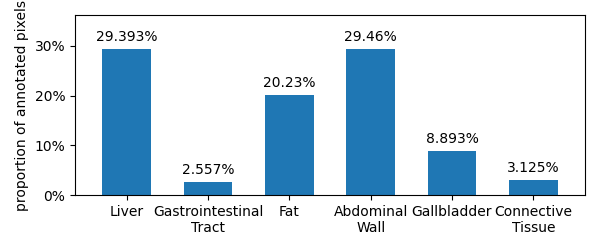}}
  
    ~
    
    \subfloat[ht][Group with smaller proportion]{
        \includegraphics[width=0.9\columnwidth]{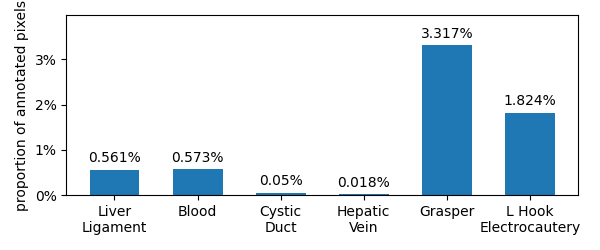}}
    
    ~
    
    \caption{Proportion of annotated pixels (y-axis) per class
      (x-axis)}
	\label{fig:proportion_annote}
\end{figure}

In this dataset, not all 13 classes appear in every frame at
the same time. The ratios of annotated pixels of 13 classes are
shown in Figure~\ref{fig:proportion_annote}. As one may notice, the
classes are not well balanced, which may lead to poor training results
if using without cautions.
Figure~\ref{fig:proportion_annote}(a) shows the annotated ratio of the
classes whose ratio of annotated pixels are greater than $1\%$. Among them, the liver class has almost $30\%$ annotated
pixels and gallbladder class has $10\%$ annotated pixels among total
annotated pixels. Abdominal wall class, which often appears in the
background of the image, also has $30\%$ annotated pixels among total
annotated pixels. The gastrointestinal tract tissues, including
stomach and small intestine, are located next to the liver and gallbladder and,
hence, are frequently viewable. Its annotated ratio is greater than $1\%$
as well.  Figure~\ref{fig:proportion_annote}(b) shows the annotated
ratio of the classes whose ratio of annotated pixels are less than
$1\%$. The L hook electrocautery and grasper classes are two
frequently used tools in cholecystectomy surgeries. However, their
physical sizes are relatively small, compared to other objects in the
images. Hence, the ratio of their annotated pixels are less than
$1\%$.

In the dataset, each frame has three corresponding masks, including
one color mask, one annotation mask used by annotation tool, and one
watershed mask. The annotated classes for different objects are
presented in both color mask and watershed mask. The color mask is
annotated for visualization purpose, as shown in
Figure~\ref{fig:ExampleEndoscopeImage}(b) as an example. The watershed
mask is presented for the sake of programming, where each annotated
pixel has the same class ID value for three color channels.
PixelAnnotaionTool~\cite{Breheret:2017} is used to annotate the images
and generate three masks in PNG format. The annotation tool allows the
users to annotate the pixels for pre-defined classes and generate the
color mask and the watershed mask.

\begin{figure}[htbp]
	\centering
        \includegraphics[width=0.6\columnwidth]{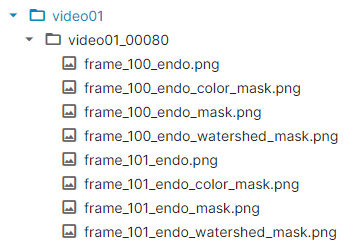}
    \caption{Example of annotation masks and raw images}
	\label{fig:mask}
\end{figure}

Figure~\ref{fig:mask} shows the example masks and raw images for {\tt
  video01\_00080} folder.
Each item in the dataset consists of four files: raw image, color
mask, annotation mask, and watershed mask. In CholecSeg8K dataset,
four files are all stored in PNG format. 

%% file: 03Example.tex
\section{Examples of Annotated Data}
\begin{figure*}[ht]
  \centering
    \subfloat[Example Endoscope Image of Gallbladder]{
        \includegraphics[width=1\columnwidth]{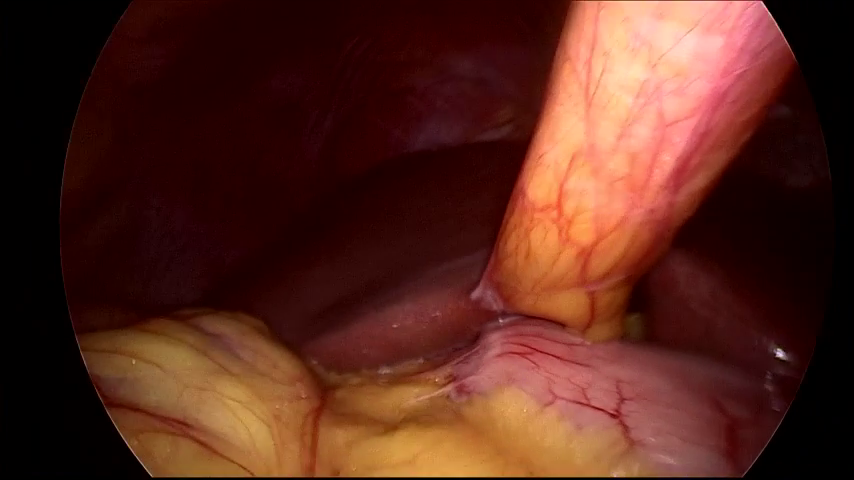}}
    ~
    \subfloat[Example Color Mask of Gallbladder and nearby tissues]{
        \includegraphics[width=1\columnwidth]{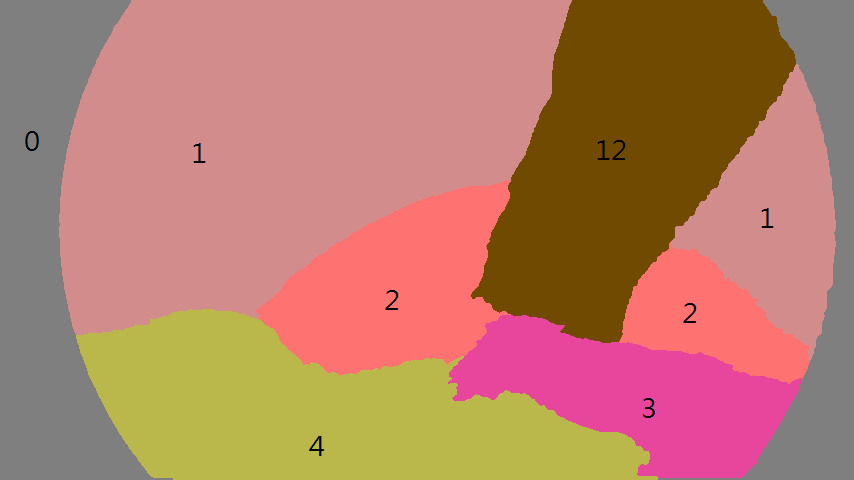}}
    ~
    \caption{Example of Semantic Segmentation Annotation of Gallbladder Endoscope Image}
    \label{fig:ExampleEndoscopeImage_1}
\end{figure*}

\begin{figure*}[ht]
	\centering
    \subfloat[Example Endoscope Image of Liver and Gallbladder]{
        \includegraphics[width=1\columnwidth]{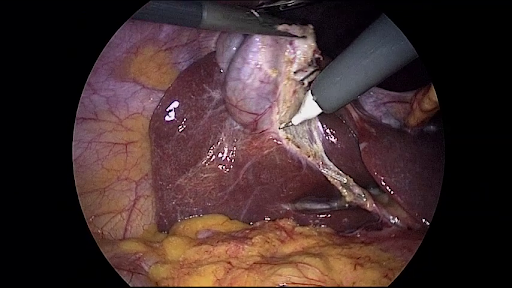}}
    ~
    \subfloat[Example Color Mask of Liver and Gallbladder]{
        \includegraphics[width=1\columnwidth]{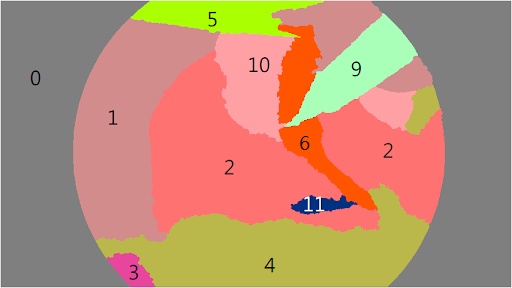}}
    ~
    \caption{Example of Semantic Segmentation Annotation for Liver and
      Gallbladder Endoscope Image} 
	\label{fig:ExampleEndoscopeImage_2}
\end{figure*}

\begin{figure*}[htbp]
	\centering
    \subfloat[Example Endoscope Image of Liver and Gallbladder]{
        \includegraphics[width=1\columnwidth]{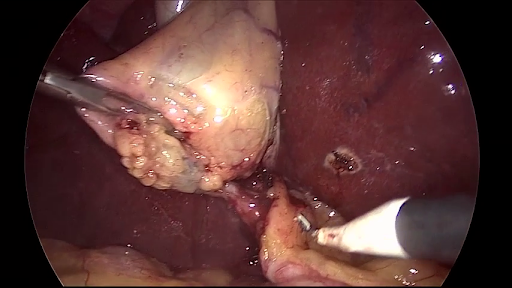}}
    ~
    \subfloat[Example Color mask of Liver and Gallbladder]{
        \includegraphics[width=1\columnwidth]{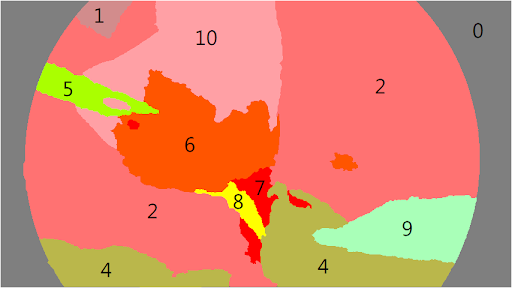}}
    ~
    \caption{Example of Semantic Segmentation Annotation of  Liver and Gallbladder in Endoscope Image}
	\label{fig:ExampleEndoscopeImage_3}
\end{figure*}

This section presents several pairs of the extracted images and color
masks, i.e., annotations of class IDs, from the dataset as examples to
demonstate the dataset. Figure~\ref{fig:ExampleEndoscopeImage_1},~\ref{fig:ExampleEndoscopeImage_2},
and~\ref{fig:ExampleEndoscopeImage_3} are three representative
examples. The left-hand side of the figures show the raw images extracted
from Cholec80 dataset and the right-hand side of the figures show the
color mask of the raw images in CholecSec8K dataset.

Figure~\ref{fig:ExampleEndoscopeImage_1} is a simple case and shows
the gallbladder (class ID 10), ligamant (class ID 12), and fat (class
ID 4) in endoscope image. Part of the liver (class ID is 2) is also
viewable. The tissues annotated as ID 12 is the round ligament of
liver. As mentioned earlier, we do not specifically annotate different
ligament of liver in this dataset and use only one class for all liver
ligaments.


Figure~\ref{fig:ExampleEndoscopeImage_2} shows a more complicated case
which has 11 classes and instruments in the image. The image contains
the liver (class ID 2) in the center and gallbladder (class ID 10) at
top central of the image.  Two instruments, grasper (class ID 5) and
L-Hook electrocautery (class ID 10), are also viewable and annotated.
Due to the low brightness in the image, the annotation at the edge of
the FoV remains non-trivial for human experts. For example, in this
figure, the top edge of grasper (class ID 5) is not trivial to
identify. Such ambiguous cases are left as it is because the dataset
aims for segmenting human organs. 

Figure~\ref{fig:ExampleEndoscopeImage_3} shows the liver and
gallbladder from a different angle.
This example also contains the pixels of 11 classes. 

%% file: 04Access.tex
\section{Access to the dataset}
\begin{figure*}[htbp]
  \centering
  \includegraphics[width=1.6\columnwidth]{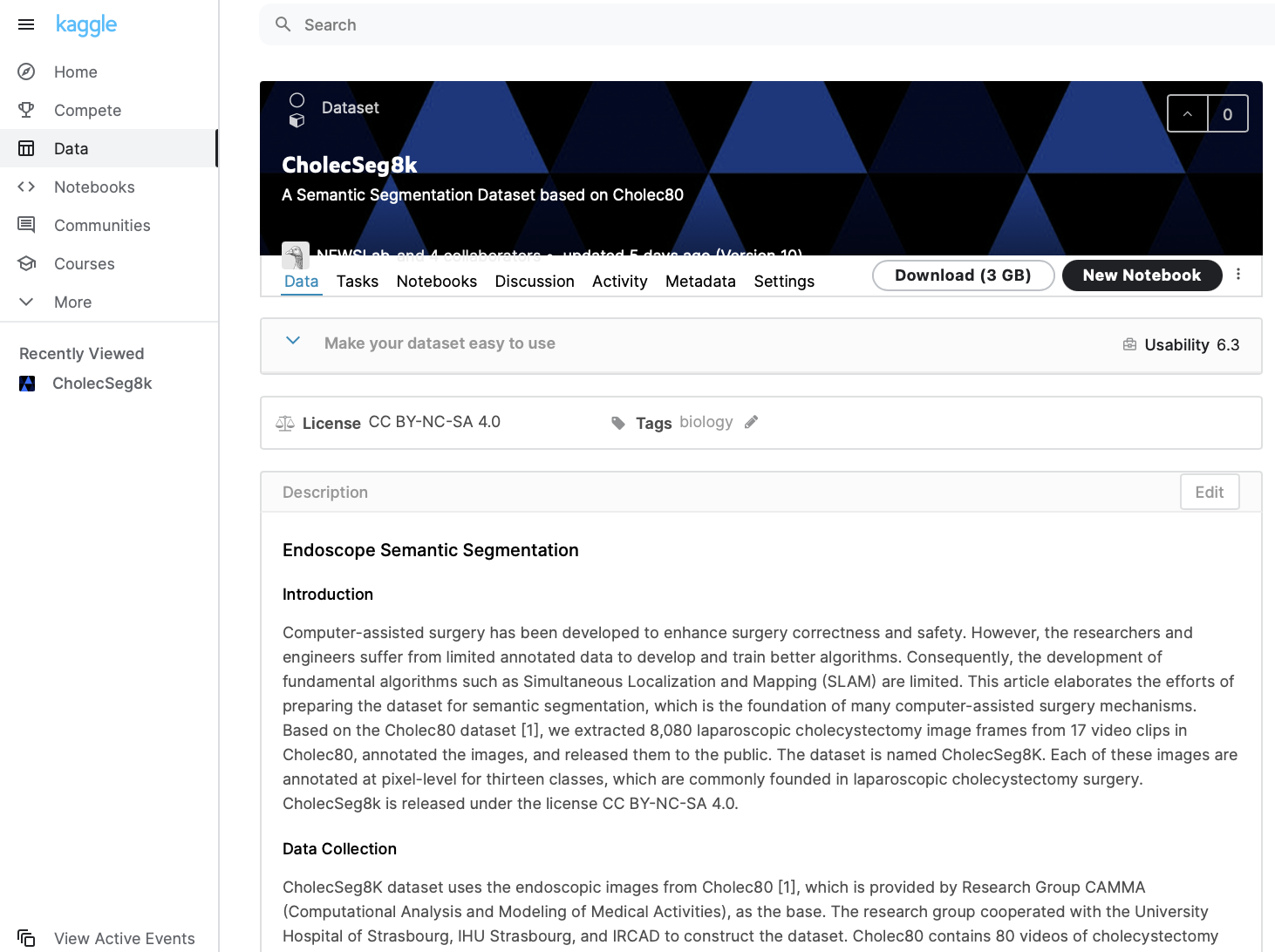}
  \caption{Screenshot of CholecSeg8K dataset page on www.kaggle.com}
  \label{fig:KagglePage}
\end{figure*}

The CholecSeg8k dataset is released under the license CC BY-NC-SA 4.0
and freely available on the Kaggle website~\cite{CholecSeg8k}.
Figure~\ref{fig:KagglePage} shows the screenshot of the CholecSec8K
data-set page on \url{www.kaggle.com}.
The instructions on using the dataset is available on the dataset page.

\section*{Acknowledgement}
This research was supported in part by the Ministry of Science and
Technology of Taiwan (MOST 108-2218-E-002-020).
\balance